\documentclass[10pt, a4paper]{article}
\usepackage{lrec-coling2024}
\usepackage{times}
\usepackage{url}
\usepackage{latexsym}
\usepackage{textalpha}
\usepackage[T1]{fontenc}
\usepackage[utf8]{inputenc}
\usepackage{amsmath}
\usepackage{booktabs}
\usepackage{graphicx}
\usepackage{enumitem}
\usepackage{multirow}
\usepackage{comment}

\title{GPT-SW3: An Autoregressive Language Model for the Scandinavian Languages}

\name{Ariel Ekgren$^1$$^\star$, Amaru Cuba Gyllensten$^1$, Felix Stollenwerk$^1$, Joey Öhman$^1$, \\
{\bf \large{Tim Isbister$^1$, Evangelia Gogoulou$^2$, Fredrik Carlsson$^2$, Judit Casademont$^1$,}} \\
{\bf \large{ Magnus Sahlgren$^1$}}}

\address{$^1$AI Sweden, Sweden \\
   $^2$RISE, Sweden \\
  $^\star$Corresponding author: \texttt{ariel.ekgren@ai.se}}

\abstract{
This paper details the process of developing the first native large generative language model for the North Germanic languages, GPT-SW3. 
We cover all parts of the development process, from data collection and processing, training configuration and instruction finetuning, to evaluation, applications, and considerations for release strategies. We discuss pros and cons of developing large language models for smaller languages and in relatively peripheral regions of the globe, and we hope that this paper can serve as a guide and reference for other researchers that undertake the development of large generative models for smaller languages.
 \\ \newline  }

\begin{document}

\maketitleabstract

\section{Introduction}

There is a growing interest in building and applying Large Language Models (LLMs) for languages other than English. This interest has been fuelled partly by the unprecedented popularity of ChatGPT\footnote{\url{chat.openai.com}} that has propelled LLMs to the forefront of general awareness, and partly by the rapid commoditization of frameworks and infrastructure for training LLMs, which has drastically lowered the threshold for researchers to train and utilize LLMs. However, even with the existence of accessible frameworks such as Hugging Face Transformers\footnote{\url{huggingface.co/docs/transformers}} and commoditized compute infrastructure either through cloud or various national (and international) supercomputer initiatives, there are significant challenges to develop LLMs for smaller languages.

The perhaps most obvious challenge is access to sufficient amounts of diverse, high-quality data. Apart from the basic question whether sufficient amounts of data at all exists for a smaller language, there may be additional complicating issues related to compliance with regulatory frameworks such as GDPR, the EU AI Act, and questions pertaining to copyright. We describe our data collection efforts in Section \ref{sec:data} (and in a separate paper, \citet{ohman2023nordic}). 
Another challenge for prospective developers of LLMs is access to sufficient amounts of compute. Some countries have national compute infrastructure devoted to researchers, but such infrastructure may have limited GPU-resources, and access is typically regulated via specific allocation tiers, which may not be suitable for large-scale projects such as LLM training. On the other hand, cloud providers are typically always an easily accessible option, but can be prohibitively costly.

We have faced all of these challenges in our work on developing the first native LLM for the Scandinavian (or, more accurately, {\it North Germanic}) languages. The LLM, which we call {\bf GPT-SW3}, is a continuation of our previous Swedish-only model \cite{ekgren-etal-2022-lessons}. GPT-SW3 is a collection of large decoder-only pretrained Transformer language models trained with a causal language modeling objective on a dataset containing approximately 320B tokens in Swedish, Norwegian, Danish, Icelandic, and English, as well as a set of 4 programming languages (Python, JavaScript, SQL and Shell script). The 
suite of models ranges from 126M to 40B parameters, and instruction-tuned versions are also available for some of these models. This paper details the entire development process, from data collection and processing, training configuration and instruction-tuning, to evaluation and considerations for model release. 

\section{Related Work}
\label{sec:related}

\begin{table*}[!ht]
    \centering
    \footnotesize
    \begin{tabular}{r|l|l|l|l}
    \toprule
        Size & Language & Open & Name & Reference \\
        \midrule
        20B & English & Yes & GPT-NeoX & \citet{gptneox_2022} \\
        30B & English & Yes & MPT & \citet{MosaicML2023Introducing} \\
        34B & Chinese, English & Yes & Yi & \citet{ai2024yi} \\
        34B & Finnish, English & Yes & Poro & \url{huggingface.co/LumiOpen/Poro-34B} \\
        \midrule
        \multirow{3}{*}{{\bf 40B}} & {\bf Swedish, Norwegian} & \multirow{3}{*}{{\bf Yes}} & \multirow{3}{*}{{\bf GPT-SW3}} & \multirow{3}{*}{{\bf This paper}} \\ 
        & {\bf Danish, Icelandic} & & & \\
        & {\bf Faroese, English} & & & \\
        \midrule
        45B & Multilingual & Yes & Mixtral & \citet{jiang2024mixtral} \\
        50B & English & No & BloombergGPT & \citet{wu2023bloomberggpt} \\
        65B & Multilingual & Yes & LLaMA & \citet{touvron2023llama} \\
        70B & Multilingual & Yes & LLaMA-2 & \citet{touvron2023llama2} \\
        70B & English & No & Chinchilla & \citet{hoffmann2022an} \\
        72B & Chinese, English & Yes & Qwen & \citet{bai2023qwen} \\
        100B & Russian & Yes & YaLM & \url{github.com/yandex/YaLM-100B} \\
        120B & English & No & Galactica & \citet{galactica} \\
        130B & Chinese, English & Yes & GLM & \citet{glm} \\
        137B & English & No & LaMDA & \citet{lamda} \\
        175B & English & No & GPT-3 & \citet{gpt3_2020} \\
        175B & English & Yes & OPT & \citet{opt} \\
        176B & Multilingual & Yes & BLOOM & \citet{bloom_model_2022} \\
        178B & English & No & Jurassic-1 & \citet{J1WhitePaper} \\
        180B & Multilingual & Yes & Falcon & \citet{falcon40b} \\
        200B & Chinese & No & PanGu-$\alpha$ & \citet{pangu} \\
        260B & Chinese & No & Ernie 3.0 & \citet{ernie3} \\
        280B & English & No & Gopher & \citet{gopher_2021} \\
        314B & Multilingual & Yes & Grok-1 & \url{github.com/xai-org/grok-1} \\
        530B & English & No & Megatron-Turing & \citet{megatron-turing} \\
        540B & English & No & PaLM & \citet{palm_2022} \\
        \midrule
        ? & Multilingual & No & Mistral & \url{chat.mistral.ai} \\
        ? & Multilingual & No & Gemini & \url{gemini.google.com} \\
        ? & Multilingual & No & GPT-4 & \url{openai.com/gpt-4} \\
        ? & Multilingual & No & Claude & \url{anthropic.com/claude} \\
        \bottomrule
    \end{tabular}
    \caption{LLMs with a parameter count of more than 20 billion, sorted by ascending size. The ``language'' column indicates the languages in the pretraining data (excluding code, which is present in a majority of models), and the ``open'' column indicates whether the model weights are accessible for download.}
    \label{tab:model_sizes}
\end{table*}

The era of LLMs arguably started with the 175B parameter GPT-3 model that was introduced in 2020 \cite{gpt3_2020}. During the last 3 years, we have seen a steady stream of new LLMs, exemplified by the models listed in Table \ref{tab:model_sizes}. What counts as a ``large'' language model is of course not rigorously defined. Our compilation in Table \ref{tab:model_sizes} lists models that have been trained from scratch with more than 20 billion parameters, but this threshold is arbitrary, and could as well be 1B or 100B. 

The majority of current LLMs are built from English data. There are however a growing number of exceptions to this, including a number of Chinese models (Yi, Qwen, GLM, PanGu-$\alpha$, and Ernie 3.0), one Russian (YaLM) and one Finnish/English model (Poro), as well as a number of models that count as multilingual since they include a number of different languages; examples include Falcon, Mixtral, LLaMA (1 and 2), BLOOM, Grok-1 as well as the commercial models Mistral, Gemini, GPT-4 and Claude.
GPT-SW3 is unique in the sense that it is the only model trained specifically on the North Germanic languages (Swedish, Norwegian, Danish, Icelandic and Faroese), and as such it is also the only current LLM built to represent a specific language group. 

Most of the early LLMs from 2020 and 2021, such as GPT-3, Jurassic-1, PanGu-$\alpha$, Ernie 3.0 and Gopher, were not (and are still not) publicly released. However, from 2022 onwards there has been a noticeable shift in release strategies from developers of LLMs, with an increasing number of models being released under various forms of more or less permissive licenses that allow for downloading of model weights (what we refer to as ``open'' in Table \ref{tab:model_sizes}). GPT-SW3 is also publicly released under a permissive license.   
Section \ref{sec:release} provides a more detailed discussion about our considerations regarding release strategies.

\section{Data}
\label{sec:data}

The arguably most challenging aspect of building an LLM for a (set of) smaller languages is finding sufficient amounts of text data with sufficient quality and variety. Since there are no readily available large data collections for LLM pretraining in the North Germanic languages, we compiled our own training data, which we call {\em The Nordic Pile} \cite{ohman2023nordic}. 

\begin{table*}[]
\scriptsize
\begin{center}
\begin{tabular}{@{}l|rrrrrrr|r@{}}
\toprule
               & Swedish   & English   & Norwegian & Danish    & Icelandic & Other   & Code     & Total             \\
\midrule
Articles       & 16.49 GB  & 173.52 GB & 0.01 GB   & 0.19 GB   &           &         &          & 190.21 GB         \\
Books          & 1.15 GB   & 94.14 GB  & 0.04 GB   & 0.06 GB   &           &         &          & 95.39 GB          \\
Conversational & 65.61 GB  & 81.67 GB  & 0.57 GB   & 2.84 GB   & 0.07 GB   & 0.01 GB &          & 150.77 GB         \\
Math           & 4.58 GB   & 4.98 GB   & 0.01 GB   & 0.01 GB   &           & 0.19 GB &          & 9.77 GB           \\
Miscellaneous  & 28.85 GB  & 56.31 GB  & 48.48 GB  & 13.85 GB  & 10.26 GB  & 1.8 GB  &          & 159.55 GB         \\
Web CC         & 188.94 GB & 60.36 GB  & 90 GB     & 111.33 GB & 8.79 GB   & 2.05 GB &          & 461.47 GB         \\
Web Sources    & 7.83 GB   & 0.61 GB   & 0.03 GB   & 1.85 GB   &           &         &          & 10.32 GB          \\
Wikipedia      & 1.03 GB   & 14.77 GB  & 0.48 GB   & 0.38 GB   & 0.05 GB   &         &          & 16.71 GB          \\
Code           &           &           &           &           &           &         & 114.5 GB & 114.5 GB          \\
\midrule
Total          & 314.48 GB & 486.36 GB & 139.62 GB & 130.51 GB & 19.17 GB  & 4.05 GB & 114.5 GB & 1,208.69 GB        \\
\bottomrule
\end{tabular}
\caption{Data sizes for each language and category after cleaning and processing.}
\label{tab:category_language_sizes}
\end{center}
\end{table*}

Our training data consists of text data collected from various open general data sources, such as MC4 \cite{mt5_mc4_2021}, OSCAR \cite{oscar_2019,oscar_2020}, OPUS \cite{opus_2004}, Wikipedia and The Pile \cite{pile_2021}, as well as language-specific corpora such as the Norwegian Colossal Corpus \cite{kummervold-etal-2021-operationalizing}, the Danish and Icelandic Gigaword corpora \cite{stromberg-derczynski-etal-2021-danish,barkarson-etal-2022-evolving}, and various data repositories, websites and discussion forums in Swedish. We also include a set of four different programming languages from the CodeParrot\footnote{\url{huggingface.co/codeparrot}} collection (Python, JavaScript, SQL and Shell script). Table \ref{tab:category_language_sizes} summarizes the various data categories across the various languages included in the training data. 

We performed several steps of data processing on the collected data, including normalization, quality filtering and deduplication (both exact and fuzzy). The normalization takes care of non-printing characters, and normalizes whitespace and Unicode characters. The quality filtering applies a set of heuristics inspired by Gopher and ROOTS  \cite{gopher_2021,bloom_data_2022}, and the fuzzy deduplication utilizes MinHash LSH \cite{minhash_1997}.
Our training data, processing steps, and arguments for selection and filtering of sources is described in more detail in a separate publication \cite{ohman2023nordic}.

We weight the different languages and categories (cf.~Table \ref{tab:category_language_sizes}) in such a way that the composition of the training data changes, while its total size stays the same. Note that this implies that some data are used multiple times while other data are discarded. More details can be found in App.~\ref{app:weighting}. 
After weighting, we end up with the following distribution of data in terms of languages:
\begin{itemize}
    \item Swedish: $35.3\%$ 
    \item English: $23.4\%$ 
    \item Norwegian: $17.3\%$
    \item Danish: $14.8\%$
    \item Icelandic: $2.7\%$ 
    \item Code: $6.5\%$
\end{itemize}

\section{Tokenizer}
\label{sec:tokenizer}

We employed the SentencePiece library \cite{kudo-richardson-2018-sentencepiece} to train a Byte-Pair Encoding \cite{sennrich2016neural} tokenizer on a representative 1\% sample of the model training data. The tokenizer has a vocabulary size of 64,000. Our reason for using a slightly larger vocabulary size compared to other LLMs (e.g.~GPT-3, OPT, and GPT-NeoX have a vocabulary size of 50k tokens, while LLaMA only uses 32k tokens in the vocabulary) is that we want to improve the performance of the smaller languages included in our data, such as Icelandic.

Our tokenizer works without explicit pretokenization. However, it splits digits and uses SentencePiece's dummy prefix and the byte fallback feature. We also added repeated whitespace tokens \cite{gptneox_2022} and special code tokens like $\textlangle \textbar python\textbar \textrangle$ to the tokenizer's vocabulary, in order to improve the way code data is handled. Note that the special code tokens are present in the code data as well. 
We describe the tokenizer's features, training and evaluation in more detail in a separate paper \cite{gptsw3_tokenizer}.

After tokenization, our training data consists of around 320B tokens.

\section{Training}
\label{sec:training}

We trained our models on 160 40GB A100 GPUs
using the Nemo Megatron framework \cite{narayanan2021efficient}. 
We trained models of increasing size, starting out with the smaller models. This strategy was employed to identify problems with the pretraining procedure early on, before training the larger models.\footnote{One such problem did occur: In the initial runs of the small models, we assumed that the tokenization and binarization script would delimit documents by end-of-text-tokens. It did no such thing, which lead to frequent and unexpected context-switching during generation. Thankfully, this was discovered and remedied early on due to the training strategy.} 

Table~\ref{tab:training_runs} shows the most important hyperparameters for the various model sizes. All models have the same vocabulary size (64,000) and sequence length (2,048). The feed-forward dimension is always four times the embedding dimension. The models were trained using packing, meaning that each sample in a batch can consist of multiple documents\footnote{We have $\sim$4 documents per sample on average.} delimited by end-of-text-tokens\footnote{Due to the nature of dataloading in Nemo Megatron, this also means that documents belonging to the same data sample also belong to the same dataset. Combined with cross-document attention, this could have a slightly adverse effect on the end-of-text token as documents across end-of-text boundaries are not completely independent.}. We did \emph{not} use attention-masking between documents. 

\begin{table}[]
\centering
\scriptsize
\begin{tabular}{c|ccccc}
\toprule
Size & lr & batch & heads & depth & emb. dim. \\ 
\midrule
126M & 3e-4 & 256 & 12 & 12 & 768  \\
356M & 3e-4 & 256 & 16 & 24 & 1,024 \\
1.3B & 2e-4 & 512 & 32 & 24 & 2,048 \\
6.7B & 1.2e-4 & 1,000 & 32 & 32 & 4,096 \\
20B & 1.4e-4 & 1,920 & 48 & 44 & 6,144 \\
40B & 1.1e-4 & 1,920 & 64 & 48 & 8,192 \\
\bottomrule 
\end{tabular}
\caption{Hyperparameters used for the GPT-SW3 models of different sizes. All models have the same vocabulary size (64,000) and sequence length (2,048). The number of model parameters are denoted by Size, while lr corresponds to the maximum learning rate.}
\label{tab:training_runs}
\end{table}
\begin{figure}[]
\includegraphics[width=\linewidth]{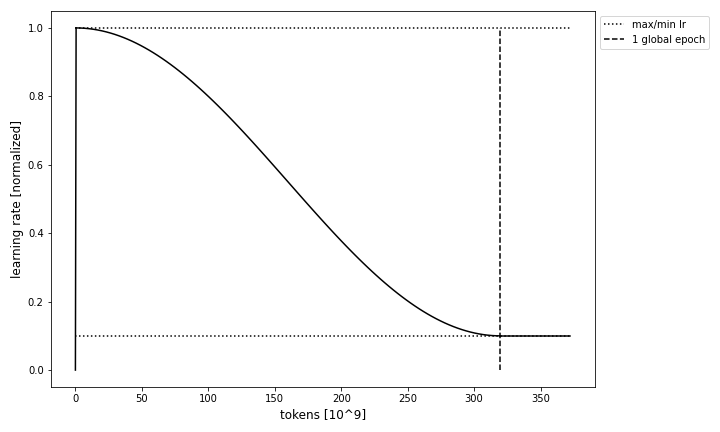}
\centering
\caption{Normalized learning rate schedule. The maxima of the learning rate are given in Table \ref{tab:training_runs}.}
\label{fig:scaling_strategy_learning_rate_schedule}
\end{figure}
The learning rate schedule we employed is a function of the amount of data, as visualized in Figure \ref{fig:scaling_strategy_learning_rate_schedule}.
It is the same for all model sizes apart from a global factor, the maximum learning rate listed in Table \ref{tab:training_runs}.
The training process starts with a short warm-up period that amounts to 0.5B tokens, during which the learning rate is increased from 0 to its maximum. Afterwards, we use a cosine decay to the minimum learning rate ($\frac{1}{10}$ of the maximum learning rate) for another 319.3B tokens. Finally, training is continued at a constant learning rate until up to 372.2B tokens are reached.

Table \ref{tab:single_run_flop} shows the model FLOP/s we achieved for the various model sizes, as well as the utilization w.r.t.~peak theoretical FLOP/s. For the larger models, the utilization numbers are comparable to those reported by NVIDIA in \citet{narayanan2021efficient} whereas the smaller models show much poorer utilization. We believe this can be attributed to over-parallelization of the smaller models.

\begin{table}[t!]
    \centering
    \scriptsize
    \begin{tabular}{c|ccc}
    \toprule
    Size &  GPUs & FLOP/s & Utilization \\
    \midrule 
    126M &  64 & $1.71 \times 10^{15}$ & 8.58\% \\
    356M &  32 & $1.57 \times 10^{15}$ & 15.69\%\\
    1.3B  & 128 & $5.21 \times 10^{15}$ & 13.06\% \\
    6.7B &  160 & $8.63 \times 10^{15}$ & 17.28\% \\
    20B & 160 & $1.89 \times 10^{16}$ & 37.91\% \\
    40B & 160  & $1.96 \times 10^{16}$ & 39.23\% \\
    \bottomrule
    \end{tabular}
    \caption{Achieved Model FLOP/s and Utilization (w.r.t.~peak theoretical FLOP/s) for the various model sizes during a single job.}
    \label{tab:single_run_flop}
\end{table}

\begin{figure}
    \centering
    \includegraphics[scale=0.07]{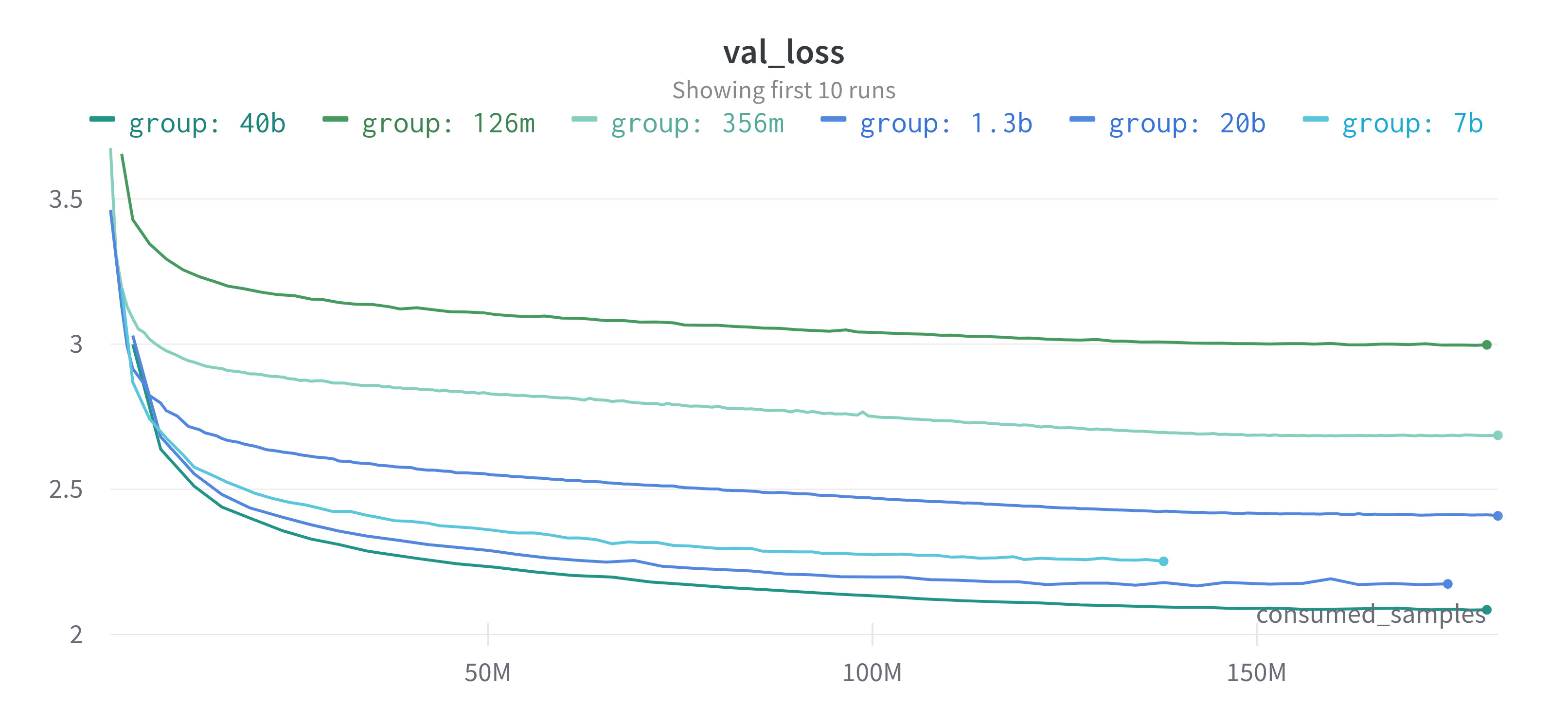}
    \caption{Validation loss during training.}
    \label{fig:val_loss}
\end{figure}

The validation loss during training can be seen in Figure \ref{fig:val_loss}. As expected, the larger models reach lower validation loss, and largely follow the expected scaling behavior (see Appendix \ref{app:scaling} for more details). 
Contrary to some other works \cite{opt}, we did not experience any divergence during training, but we did observe the occasional gradient spike (with no catastrophic long-term effect). 

\subsection{Energy consumption}

We estimate that the total compute budget for our 
training runs is something like 560k GPU hours. This is obviously a very rough estimate, and likely to be somewhat on the high end. The average carbon intensity in Sweden, where the models were trained, is estimated to be around 10 grams of carbon dioxide per kilowatt-hour (gCO$_2$/KWh).\footnote{\url{https://bit.ly/4anuJyK}}
Using the \texttt{ML CO$_2$ IMPACT calculator}\footnote{\url{mlco2.github.io/}} we calculate our total carbon emissions to be roughly 14,462 kg CO$_2$. This is approximately on the same level of carbon emissions as is generated by manufacturing {\em one} 80 kWh lithium-ion battery used in electric cars.\footnote{\url{https://bit.ly/3Ts9rJi}}
Considering the relatively low carbon intensity of the power system used to train our models, our carbon footprint is significantly lower than that of other similar models (see e.g.~the emissions estimated for the BLOOM model \cite{luccioni2022estimating}).

\section{Instruction finetuning}

Due to the popularity and effectiveness of instruction-tuned models such as ChatGPT, we also produce a set of instruction-tuned models. The models were fine-tuned using instruction tuning \citep{ouyang2022training} data from multiple sources: Open Assistant\footnote{\url{huggingface.co/datasets/OpenAssistant/oasst1}} \citep{köpf2023openassistant}, The Open Instruction Generalist (OIG) dataset\footnote{\url{laion.ai/blog/oig-dataset/}}, Dolly\footnote{\url{huggingface.co/datasets/databricks/databricks-dolly-15k}}, and a dataset compiled specifically for this study based on FASS (The Swedish pharmaceutical formulary, \emph{Farmaceutiska Specialiteter i Sverige}).

For the OIG data, we selected high-quality subsets, which encompassed a wide range of topics and dialog styles. The datasets selected were abstract infill, HC3 human, SODA dialog, CHIP2, image prompts instructions, SQLv1, conversation FinQA, MathQA FLANv2 Kojma COT, SQLv2, CUAD, NI, SQuAD v2, essays, OpenAI Summarize TLDR, SQuAD v2 more negative, grade school math instructions, Rallio SODA upgraded 2,048, and unnatural instructions. The rest of the datasets from OIG were discarded, leaving us with a considerably smaller data set than the original OIG.

We formatted the instruction data into a unified turn-based format, where an initial user query is followed by an assistant response, which in turn is (potentially) followed by a follow up user query, and so on.
This turn-based query-response format was formatted in two ways\footnote{This means that each conversation occurs twice in the final training data, once as chat and once in the unrolled format.}:
\begin{itemize}
    \item An unrolled format, where the query-response-query-turns are simply delimited by double newlines:
    \begin{small}
    \begin{verbatim}
        Query
        
        Response
        
        ...
    \end{verbatim}
    \end{small}
    \item An explicit chat format inspired by the chatml format:\footnote{\url{github.com/openai/openai-python/blob/main/chatml.md}}
    \begin{small}
    \begin{verbatim}
        <eos><bos>User: Query
        <bos>Assistant: Response
        <bos>...
    \end{verbatim}
    \end{small}
    Where \verb|<eos>| is the special document-delimiter token used during pretraining, and \verb|<bos>| is a special turn-delimiter token only used during instruction-finetuning.
\end{itemize}

In both cases, we employed stochastic merging of independent conversations using a simple concatenation strategy: Sample a \emph{number of samples} $N$ according to a geometric distribution, randomly sample $N$ conversations from the dataset, and let the concatenation of these $N$ conversations be your new datapoint. This was done to improve the models context-switching capabilities. 

To accommodate our multilingual focus, the OpenAssistant and Dolly data sets were translated. The OpenAssistant data set was translated from English to Swedish, Danish, Norwegian, and Icelandic, while the Dolly data set was translated from English to Swedish and Danish. The translations were done with the GPT-SW3 base models.

\begin{table*}[!h]
\begin{center}
\scriptsize
\begin{tabular}{l|cccccc|ccccccc}
\toprule
                          & \multicolumn{6}{c|}{0-shot} & \multicolumn{6}{c}{5-shot} \\
                Task      & 126M  & 356M  & 1.3B  & 6.7B  & 20B & 40B & 126M  & 356M  & 1.3B  & 6.7B  & 20B & 40B \\
\midrule
ANLI Round 1     & 0.336 & 0.298 & 0.315 & 0.337 & 0.322 & \textbf{0.368} & 0.334 & 0.313 & 0.332 & 0.317 & 0.330 & {\bf 0.350} \\
ANLI Round 2     & 0.317 & 0.338 & 0.345 & 0.333 & 0.343 & \textbf{0.358} & 0.341 & {\bf 0.359} & 0.340 & 0.332 & 0.333 & 0.350 \\
ANLI Round 3     & 0.322 & 0.325 & 0.311 & 0.332 & 0.333 & {\bf 0.391} & 0.330 & 0.319 & 0.336 & 0.330 & 0.343 & \textbf{0.364} \\
WSC              & 0.365 & 0.365 & 0.365 & 0.413 & 0.394 & {\bf 0.548} & 0.365 & 0.365 & 0.394 & 0.365 & \textbf{0.519} & 0.423 \\
HellaSwag        & 0.279 & 0.322 & 0.393 & 0.457 & 0.502 & {\bf 0.532} & 0.280 & 0.321 & 0.387 & 0.452 & 0.504 & {\bf 0.532} \\
Winogrande       & 0.493 & 0.517 & 0.571 & 0.617 & 0.632 & {\bf 0.656} & 0.522 & 0.527 & 0.557 & 0.607 & 0.657 & {\bf 0.674} \\
PIQA             & 0.584 & 0.642 & 0.707 & 0.735 & 0.768 & {\bf 0.772} & 0.600 & 0.646 & 0.708 & 0.739 & 0.765 & {\bf 0.776} \\
ARC (Easy)       & 0.388 & 0.408 & 0.549 & 0.609 & \textbf{0.692} & 0.687 & 0.412 & 0.473 & 0.588 & 0.647 & 0.707 & {\bf 0.718} \\
ARC (Cha.)       & 0.204 & 0.200 & 0.253 & 0.294 & 0.352 & {\bf 0.368} & 0.191 & 0.213 & 0.276 & 0.312 & 0.360 & {\bf 0.387} \\
OpenBookQA       & 0.140 & 0.186 & 0.214 & 0.220 & 0.268 & {\bf 0.274} & 0.148 & 0.188 & 0.228 & 0.260 & 0.240 & {\bf 0.300} \\
HeadQA           & 0.224 & 0.236 & 0.266 & 0.278 & {\bf 0.308} & 0.302 & 0.227 & 0.243 & 0.273 & 0.295 & 0.233 & \textbf{0.330} \\ 
\midrule
Average          & 0.332 & 0.349 & 0.390 & 0.420 & 0.447 & {\bf 0.478} & 0.341 & 0.361 & 0.402 & 0.424 & 0.454 & \textbf{0.473} \\
\bottomrule
\end{tabular}
\caption{LM Evaluation Harness accuracy scores of GPT-SW3 in 0-shot setting (left) and 5-shot setting (right). The best performance for each task and setting is marked in boldface.}
\label{tab:lm_harness_nlp_0shot}
\end{center}
\end{table*}

The fine-tuning process was applied consistently across models of different scales, including 356M, 1.3B, 6.7B, and 20B parameters. We used a sequence length of 2,048, a batch size of 160, and an initial learning rate of $2 \times 10^{-5}$, which we gradually reduced with cosine decay to a minimum of $2 \times 10^{-6}$ over the course of 2,069 global steps. This approach consumed a total of 331,040 samples, with a warm-up period spanning 360 steps.

\section{Evaluation}

Since we currently lack suitable evaluation benchmarks for generative language models in the North Germanic languages, we use language modeling perplexity on a set of held-out data to compare our models. We use character length normalization \cite{cotterell-etal-2018-languages,Mie2016Can} rather than token length for calculating perplexity formula, since token length favours tokenizers that use more tokens per sentence. We thus calculate perplexity as:

\begin{align}\label{ppl}
\begin{split}
    PPL_c(X) &=\exp\bigg\{-\frac{1}{\textbf{c}}\sum_{i=1}^{\textbf{t}}\log p(T_i|T_{<i})\bigg\}\\
    \textbf{c} &= \text{Character length of } X\\
    T &= \text{Tokenization of } X \\
    \textbf{t} &= \text{Token length of } T
\end{split}
\end{align}

Table \ref{tab:char_perplexity} shows the perplexity scores for GPT-SW3 in comparison with GPT-NeoX (20B) and the recent Falcon models (1B, 7B and 40B) which have been trained on a small amount of Swedish data (1B tokens). It is obvious, and perhaps not very surprising, that GPT-SW3 reach the lowest language modeling perplexity on Swedish, Danish and Norwegian data, and that larger models reach lower perplexity. 

\begin{table}[]
\begin{center}
\scriptsize
\begin{tabular}{l|c|c|c|c}
\midrule
Model & SE & DA & NO & EN    \\ \midrule
GPT-SW3 40B        & {\bf 1.9240} & {\bf 1.8698} & {\bf 1.9270} & 1.9660 \\
GPT-SW3 20B        & 1.9458 & 1.8932 & 1.9491 & 1.9928 \\
GPT-SW3 6.7B       & 1.9781 & 1.9229 & 1.9795 & 2.0152 \\
GPT-SW3 1.3B       & 2.0665 & 2.0192 & 2.0741 & 2.1166 \\
GPT-SW3 356M       & 2.1973 & 2.1568 & 2.2130 & 2.2477 \\
GPT-SW3 126M       & 2.3748 & 2.3455 & 2.3992 & 2.4297 \\
\midrule
GPT-NeoX 20B   & 2.3807 & 2.3378 & 2.4245 & 1.9377 \\
\midrule
Falcon 40B     & 2.0194 & 2.2379 & 2.2705 & {\bf 1.8152} \\
Falcon 7B      & 2.6546 & 2.7355 & 2.7740 & 1.8743 \\
Falcon-RW 1B   & 3.7672 & 3.7187 & 3.7806 & 1.9765 \\ \midrule
\end{tabular}
\caption{Evaluation of perplexity normalized on characters on held-out data for Swedish, Danish, Norwegian and English. The best score per language is marked in boldface.} 
\label{tab:char_perplexity}
\end{center}
\end{table}

The fact that GPT-SW3 has been trained on English data, and seems to perform well w.r.t.~language modeling perplexity, suggests that we can also take advantage of the English-language Language Model Evaluation Harness \cite{eval-harness} to benchmark our models. The LM Evaluation Harness framework contains a large number (200+) of different evaluation tasks. We select a small subset of these to benchmark our models (ANLI, WSC, HellaSwag, Winogrande, PIQA, ARC, OpenBookQA, and HeadQA). Table \ref{tab:lm_harness_nlp_0shot} shows the results of GPT-SW3 in a 0-shot setting (left side of the table) and 5-shot setting (right side of the table). Unsurprisingly, the larger models perform better, with the 40B model performing best overall. 

Table \ref{tab:lm_harness_instruct_comparison_nlp} shows a comparison between two of our base models and their respective instruction-tuned variants in both 0-shot and 5-shot setting. The general tendency is (perhaps unsurprisingly) that instruction-tuning is beneficial for the models when applied to the LM Harness tasks. The 6.7B instruction-tuned model even outperforms the 20B base model on average on LM Harness, and the 20B instruction-tuned version approaches the performance of the 40B model.

\begin{table*}[!h]
\begin{center}
\scriptsize
\begin{tabular}{l|cccc|cccc}
\toprule
                          & \multicolumn{4}{c|}{0-shot} & \multicolumn{4}{c}{5-shot} \\
                Task      & 6.7B  & 6.7B-instruct & 20B & 20B-instruct & 6.7B  & 6.7B-instruct & 20B & 20B-instruct \\
\midrule
ANLI Round 1    & \textbf{0.337} & 0.329 & 0.322 & \textbf{0.372} & \textbf{0.317} & 0.309 & 0.330 & \textbf{0.328} \\
ANLI Round 2    & 0.333 & \textbf{0.376} & 0.343 & \textbf{0.381} & 0.332 & \textbf{0.341} & 0.333 & \textbf{0.359} \\
ANLI Round 3     & 0.332 & \textbf{0.361} & 0.333 & \textbf{0.378} & 0.330 & \textbf{0.331} & 0.343 & \textbf{0.372} \\
WSC              &  \textbf{0.414} & 0.385 & \textbf{0.394} & 0.365 & 0.365 & \textbf{0.490} & \textbf{0.519} & 0.375\\
HellaSwag        & 0.457 & \textbf{0.503} & 0.502 & \textbf{0.528} & 0.452 & \textbf{0.502} & 0.504 & \textbf{0.527} \\
Winogrande       & \textbf{0.617} & \textbf{0.617} & 0.632 & \textbf{0.639} & 0.607 & \textbf{0.616} & \textbf{0.657} & 0.646 \\
PIQA             & 0.735 & \textbf{0.765} & \textbf{0.768} & 0.764 & 0.739 & \textbf{0.762} & 0.766 & \textbf{0.773} \\
ARC (Easy)       & 0.609 & \textbf{0.679} & \textbf{0.692} & 0.677 & 0.650 & \textbf{0.686} & \textbf{0.707} & 0.702 \\
ARC (Cha.)       & \textbf{0.294} & 0.352 & 0.352 & \textbf{0.355} & 0.312 & \textbf{0.360} & 0.360 & \textbf{0.382} \\
OpenBookQA       & 0.220 & \textbf{0.274} & \textbf{0.268} & 0.250 & 0.260 & \textbf{0.288} & 0.240 & \textbf{0.282} \\
HeadQA           & 0.278 & \textbf{0.318} & 0.309 & \textbf{0.310} & 0.295 & \textbf{0.334} & 0.233 & \textbf{0.323} \\
\midrule
Average          & 0.421 & {\bf 0.451} & 0.447 & {\bf 0.457} & 0.424 & {\bf 0.456} & 0.454 & \textbf{0.461} \\
\bottomrule
\end{tabular}
\caption{LM Evaluation Harness accuracy scores of our instruct models (6.7B and 20B) compared with their same-size base model counterparts in 0-shot setting (left) and 5-shot setting (right). The best performance for each task and setting is marked in boldface.}
\label{tab:lm_harness_instruct_comparison_nlp}
\end{center}
\end{table*}

\begin{table}[]
\centering
\scriptsize
\begin{tabular}{l|rrr}
\toprule
{} &  GPT-NeoX &  DaVinci &  GPT-SW3 40B \\
\midrule
ANLI Round 1     &     0.340 &    0.363 &       \textbf{0.368} \\
ANLI Round 2     &     0.343 &    \textbf{0.375} &       0.358 \\
ANLI Round 3     &     0.354 &    0.369 &       \textbf{0.391} \\
WSC              &     0.500 &    \textbf{0.548} &       \textbf{0.548} \\
HellaSwag        &     0.535 &    \textbf{0.592} &       0.532 \\
Winogrande       &     0.661 &    \textbf{0.699} &       0.656 \\
SciQ             &     0.928 &    0.949 &       \textbf{0.955} \\
PIQA             &     0.779 &    \textbf{0.791} &       0.772 \\
ARC (Easy)       &     0.723 &    \textbf{0.762} &       0.687 \\
ARC (Challenge)  &     0.380 &    \textbf{0.435} &       0.368 \\
OpenBookQA       &     0.290 &    \textbf{0.336} &       0.274 \\
LogiQA           &     0.230 &    0.227 &       \textbf{0.290} \\
PROST            &     \textbf{0.296} &    0.267 &       0.263 \\
\midrule
Average          & 0.489 & {\bf 0.516} & 0.497 \\
\bottomrule
\end{tabular}
    \caption{Comparison between GPT-NeoX 20B, OpenAIs DaVinci (175B), and GPT-SW3 40B model on LM Harness (0-shot). The best score for each task is marked in boldface.}
    \label{tab:lm_harness_comparison}
\end{table}

Table \ref{tab:lm_harness_comparison} shows a comparison between GPT-SW3 40B, GPT-NeoX (20B), and GPT-3 DaVinci (presumably 175B) on LM Harness in a 0-shot setting. The models perform more or less comparably over all tests, with DaVinci outperforming GPT-NeoX and GPT-SW3 in 8 out of 13 tests, GPT-SW3 outperforming the other in 5 tests, and GPT-NeoX outperforming the others in only one test. It should be noted that these models are strictly not comparable due to significant differences in training data and parameter count; DaVinci is by far the largest of these models with (presumably) 175B parameters compared to 40B for GPT-SW3 and 20B for GPT-NeoX. On the other hand, GPT-SW3 has been trained on significantly less English data than the other models, but still performs comparably on these tests. 

\section{Release plan}
\label{sec:release}

As we touched upon in Section \ref{sec:related}, it is not obvious that new LLMs are released openly, and different developers have opted for different release strategies. Some opt for a completely open release where the weights of the model can be downloaded freely and the user is permitted to both modify and redistribute the weights, as well as to integrate the model in various types of applications, both academic and commercial. Others opt to not share the model weights at all, due to reasons such as commercial advantage, concerns about the potential for misuse, or legal restrictions relating to, e.g., the General Data Protection Regulation (GDPR). \citet{solaiman2023gradient} provides a good discussion and overview of the complexities involved in defining a suitable release strategy for an LLM. 

As a compromise between openness and caution, our release strategy consisted of two phases:

\begin{enumerate}
    \item An initial {\em restricted pre-release}, which included manual audit of applications where access was granted to the model weights for organizations and individuals in the Nordic NLP ecosystem that aimed to use the models for research purposes. Usage was also restricted by a slightly modified version of the BigScience Responsible AI License (RAIL).\footnote{\url{bigscience.huggingface.co/blog/the-bigscience-rail-license}} Our intention was to use this pre-release phase for collecting input on model behavior, flaws and limitations in order to be able to make a more informed decision about open release. The restricted pre-release lasted approximately 6 months, and showed no significant flaws or adversarial effects.
    \item Due to the positive outcome of the restricted pre-release phase, we have now released the weights of the GPT-SW3 models openly under a slightly modified version of the Apache 2 license, which allows for modification, redistribution, research and commercialization of the model weights. We believe that an open release strategy is beneficial for value creation, transparency, democratization, and reproducibility.
\end{enumerate}

The GPT-SW3 models are available at: {\bf \texttt{huggingface.co/AI-Sweden-Models}}.

\section{Discussion}

This paper has detailed the development process for our family of North Germanic LLMs. It is at this point a perfectly reasonable question to ask why we at all should build a native LLM for a set of small languages with limited resources when the dominant LLMs of large corporations already can handle these languages in a reasonable (and often even superior) way. We have several answers to this question.

We believe that there is a desire and need for cultural and linguistic {\bf representativeness} by informed choices and processing of data sources, {\bf transparency} in all design choices and throughout the entire development process, {\bf democratizing} access to natively built LLMs by open or hosted release, and open {\bf validation} of model capacities as well as {\bf utilization} of existing national compute infrastructure. Perhaps most importantly, the main goal of the GPT-SW3 initiative has been to give the Nordic research community full access to the weights of a native-language LLM, something that is not currently possible with other existing LLMs.

Developing LLMs for smaller languages is admittedly a challenging endeavor when it comes to data availability, but it also opens up opportunities to select and process data sources in a more informed manner that is guided by considerations of both population representativeness, application domains and regulatory compliance. We have taken a first step in this direction, but our efforts have been constrained by limited funding and compute resources, something we believe is an all too common situation for many developers in small countries. Our initiative has been made possible by a national collaboration with a number of other organisations that have contributed to the development in various ways, e.g.~by providing access to large-scale compute. One major advantage of our initiative is the geographical location of the computer used to train our models, which is connected to a power grid with minimal carbon intensity. As such, our training runs have caused significantly lower carbon emissions than other similar projects that run on more carbon intensive power grids.

We concede that evaluation remains an issue for LLMs built for smaller languages such as the North Germanic ones. We are actively working on evaluation resources and process for Scandinavian LLMs, and we are also running a {\em validation project}, 
where stakeholders from all sectors of society validate the models for actual use in real-world applications, which span from relatively simple text generation tasks to more complex decision support functionalities. An important finding so far in the validation project is that there is a tangible need for LLMs that allow for the possibility to modify, finetune, and host the models locally. This will likely remain an important factor for LLM adoption, even if the available models are slightly less capable than the leading proprietary models.

We conclude this paper with a short note on risks in relation to LLMs. We think the current debate on the potential for apocalyptic risks incurred by LLMs is poorly nuanced and greatly exaggerated, leading to amplified and unnecessary polarization. We think a more realistic risk is the concentration of power and capital that will inevitably occur when only a small set of companies have the resources and abilities to develop, serve and distribute LLMs. Open and nationally driven LLM initiatives are vital counterparts to such developments. Another realistic risk is inflated expectations that may occur when models are not publicly accessible for validation, modification, and further development. Open development will serve to counteract this risk.

Our position is thus that open initiatives to develop LLMs for smaller languages are important and should be supported rather than hindered by regulation, funding sources, and infrastructure access programs.  

\section*{Acknowledgements}
The GPT-SW3 initiative has been enabled by the collaboration and support from the following organizations: RISE (collaboration on experiments, data storage and compute), NVIDIA (support with the deduplication code base and Nemo Megatron), Vinnova (funding via contracts 2019-02996, 2020-04658 and 2022-00949), WASP WARA media and language (access to Berzelius via SNIC/NAISS). The computations and data handling were enabled by resources provided by the National Academic Infrastructure for Supercomputing in Sweden (NAISS) and the Swedish National Infrastructure for Computing (SNIC) at Berzelius partially funded by the Swedish Research Council through grant agreements 2022-06725 and 2018-05973. Johan Raber at the National Supercomputer Center is acknowledged for assistance concerning technical and implementational aspects in making the code run on the Berzelius resources.

\nocite{*}
\section{Bibliographical References}\label{sec:reference}

\bibliographystyle{lrec-coling2024-natbib}
\bibliography{anthology, custom}

\clearpage

\appendix

\section{Data Weighting \label{app:weighting}}

We present some details regarding the weighting of the training data (see Sec.~\ref{sec:data}). First, a target distribution in terms of languages and categories is defined. This means that we specify the fraction of the total dataset that should correspond to a given language and category. Our choices are listed in Table \ref{tab:weighting_target_distribution}.
\begin{table*}[ht!]
\small
\begin{center}
\begin{tabular}{@{}l|rrrrrr|r@{}}
\toprule
& Swedish & English & Norwegian & Danish & Icelandic & Code & Total \\
\midrule
Articles       &  2.27 & 1.96 &      & 0.03 &      &      &  4.25 \\ 
Books          &  0.12 & 5.78 &      &      &      &      &  5.90 \\
Conversational & 10.34 & 6.19 & 0.11 & 0.49 & 0.03 &      & 17.17 \\
Math           &  1.59 & 0.70 &      &      &      &      &  2.29 \\
Miscellaneous  &  4.09 & 3.33 & 9.56 & 4.79 & 1.97 &      & 23.73 \\
Web CC         & 15.49 & 1.87 & 7.54 & 9.11 & 0.72 &      & 34.73 \\
Web Sources    &  1.10 &      &      & 0.23 &      &      &  1.34 \\ 
Wikipedia      &  0.29 & 3.59 & 0.12 & 0.10 & 0.01 &      &  4.11 \\
Code           &       &      &      &      &      & 6.48 &  6.48 \\
\midrule
Total          & 35.30 & 23.41 & 17.33 & 14.75 & 2.73 & 6.48 & 100.0 \\ 
\bottomrule
\end{tabular}
\end{center}
\caption{Target distribution in terms of languages and categories. The numbers denote the fraction of the total dataset in percent. Empty cells correspond to non-existing datasets, equivalent to 0. Compare to Table \ref{tab:category_language_sizes}.}
\label{tab:weighting_target_distribution}
\end{table*}

In order to obtain this target distribution, the datasets need to be weighted, i.e. either downsampled (in case there is more data available than wanted) or upsampled (in case there is less data available than wanted). We do this in such a way that the total amount of training data remains constant at $320$B tokens. Table \ref{tab:weighting} lists the number of epochs for each individual dataset needed to achieve these conditions.
\begin{table*}[ht!]
\small
\begin{center}
\begin{tabular}{@{}l|ccccccc}
\toprule
& Swedish & English & Norwegian & Danish & Icelandic & Code \\
\midrule
Articles       & 1.90 & 0.15 &      & 1.90 &      & \\
Books          & 2.11 & 0.84 &      &      &      & \\
Conversational & 2.11 & 0.84 & 2.11 & 2.11 & 2.11 & \\
Math           & 1.69 & 0.67 &      &      &      & \\
Miscellaneous  & 2.11 & 0.84 & 2.11 & 2.11 & 2.11 & \\
Web CC         & 1.05 & 0.42 & 1.05 & 1.05 & 1.05 & \\
Web Sources    & 1.69 &      &      & 1.69 &      & \\
Wikipedia      & 3.16 & 3.16 & 3.16 & 3.16 & 3.16 & \\
Code           &      &      &      &      &      & 0.74 \\
\bottomrule
\end{tabular}
\end{center}
\caption{Epochs needed in order to achieve the target distribution (see Table \ref{tab:weighting_target_distribution}) while keeping the total amount of data constant. Empty cells correspond to non-existing datasets, equivalent to 0.}
\label{tab:weighting}
\end{table*}
Note that English datasets are mostly downsampled, while the North Germanic languages are upsampled.

\section{Scaling Analysis \label{app:scaling}}

Scaling laws describe how the upstream or downstream performance of LLMs depend on the model size $N$ and dataset size $D$. \citet{hoffmann2022an} showed that the loss for their family of monolingual, English models can accurately be described by the functional form 
\begin{equation}
    L(N,D) = E + \frac{A}{N^\alpha} + \frac{B}{D^\beta}
    \label{eq:scaling_hoffmann}
\end{equation}
with fit parameters $E = 1.69$, $A = 406.4$, $B = 410.7$, $\alpha = 0.34$ and $\beta = 0.28$. 
This assumes that the learning rate follows the schedule depicted in Figure \ref{fig:scaling_strategy_learning_rate_schedule} for all dataset sizes $D$. For this reason, we can only apply the above scaling law for the dataset size $D \approx 320{\rm B}~\text{tokens}$ (cf.~Sec.~\ref{sec:training}). In that case, Eq.~(\ref{eq:scaling_hoffmann}) reduces to 
\begin{eqnarray}
   \widetilde L(N) 
   &:=& L(N, D=320{\rm B}) \nonumber \\
   &=& \widetilde E + \frac{A}{N^\alpha}
    \label{eq:scaling_hoffmann_D} 
\end{eqnarray}
with $\widetilde E := E + B / (320 \cdot 10^9)^\beta = 1.94$.

\begin{figure}
    \centering
    \includegraphics[scale=0.5]{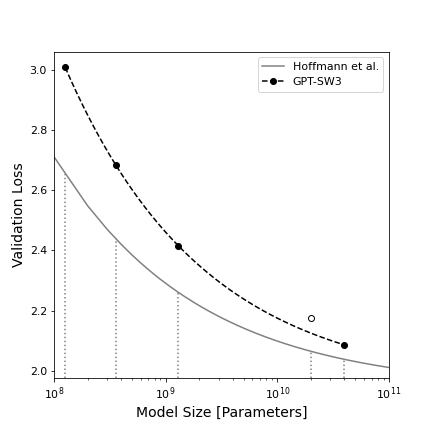}
    \caption{Scaling behaviour of GPT-SW3. The validation loss is shown as a function of the model size, while the dataset size is kept constant at 320B tokens for all models. The 20B parameter model (empty circle) is excluded from the fit (dashed curve). The gray, solid curve represents the scaling law from \citet{hoffmann2022an}.}
    \label{fig:scaling}
\end{figure}

In Figure \ref{fig:scaling}, we show the validation loss $\widetilde L(N)$ for our models as a function of the model size. 
Note that the loss for the 20B parameter model is exceptionally large. A comparison with Figure \ref{fig:val_loss} reveals that the learning curve for this very model size displays exceptional behaviour around $D \approx 320{\rm B}~\text{tokens} \approx 156{\rm M}~\text{samples}$.
We thus treat it as an anomaly and exclude it from the fit to our data.
In that case, our model's scaling behaviour can accurately be described by the functional form of Eq.~(\ref{eq:scaling_hoffmann_D}), with the fit parameters
\begin{eqnarray}
    \widetilde E_{\rm \texttt{GPT-SW3}} &=& 1.942 \pm 0.002 \\
    A_{\rm \texttt{GPT-SW3}} &=& 702.6 \pm 15.2 \\
    \alpha_{\rm \texttt{GPT-SW3}} &=& 0.348 \pm 0.001
\end{eqnarray}

Note that while $\widetilde E_{\rm \texttt{GPT-SW3}}$ and $\alpha_{\rm \texttt{GPT-SW3}}$ are very much in accordance with the results from \citet{hoffmann2022an}, $A_{\rm \texttt{GPT-SW3}} \gg A$ deviates significantly from its counterpart.
Whether this is can be attributed to our multilingual setting or has other causes is an interesting research question which we leave for future work.  

\end{document}